\documentclass[runningheads,a4paper]{llncs}
\usepackage{times}
\usepackage{graphicx}
\usepackage{latexsym}
\usepackage{amssymb}
\usepackage{amsmath}
\usepackage{amstext}
\usepackage{algorithm}
\usepackage[noend]{algorithmic}

\usepackage{color}

\begin{document}

\title{GRASP for the Coalition  Structure  Formation Problem}

\author{Nicola  Di  Mauro \and  Teresa  M.A.  Basile \and  Stefano  Ferilli  \and Floriana  Esposito
  \institute{University of Bari, Italy,\\ \{ndm,basile,ferilli,esposito\}@di.uniba.it} } 

\maketitle

\begin{abstract}
The  coalition  structure  formation problem  represents  an  active  research area  in  multi-agent
systems. A  coalition structure is defined  as a partition of  the agents involved in  a system into
disjoint coalitions.  The problem of finding the optimal coalition structure is NP-complete. 
In order to find the optimal solution in a combinatorial optimization problem it is theoretically possible to
enumerate the  solutions and  evaluate each.  But this approach  is infeasible  since the  number of
solutions often grows exponentially with the size of the problem.
In this
paper we  present a  greedy adaptive  search procedure (GRASP)  to efficiently  search the  space of
coalition  structures  in order  to  find  an optimal  one.  Experiments  and  comparisons to  other
algorithms prove the validity of the proposed method in solving this hard combinatorial problem. 
\end{abstract}

\section{Introduction}
\label{sec:introduction}

An active area of  research in multi-agent systems (MASs) is the  coalition structure formation (CSF)
of agents (equivalent to the complete set  partitioning problem). In particular it is interesting to
find coalition structures maximizing the sum of the values of the coalitions, that represent the
maximum payoff the agents belonging to the coalition can jointly receive by cooperating.
A coalition structure
(CS) is  defined as a  partition of the  agents involved in a  system into disjoint  coalitions. The
problem of finding the optimal coalition structure formation is $\mathcal{NP}$-complete.  
Coalition formation  shares a similar  structure with a  number of common problems  in theoretical
computer science, such as in combinatorial auctions  in which bidders can place bids on combinations
of items; in job shop scheduling; and as in set partitioning and set covering problems. 

Sometimes in MASs there is a time limit for  finding a solution, the agents must be reactive and they
should  act as  fast as  possible.  Hence for  the specific  task  of CSF  it is  necessary to  have
approximation algorithms  able to quickly  find solutions  that are within  a specific factor  of an
optimal solution. Hence, the goal of this paper is to propose a new algorithm for the CSF problem able to
quickly find a near optimal solution.

The problem of CSF has been studied in the context of characteristic function games (CFGs) in which the
value of each coalition is given by a characteristic function, and the values of a coalition structure
is obtained by summing the value of the contained coalitions. 
As  already  said, the  problem  of  coalition structure  formation  is  NP-hard,  indeed as  proved
in~\cite{Sandholm99}, given $n$ the number of agents, the
number   of   possible   coalition   structures    than   can   be   generated   is   $O(n^n)$   and
$\omega(n^{n/2})$. Moreover,  in order to establish any  bound from the optimal,  any algorithm must
search at least $2^n-1$ coalition structures.  

The coalition formation process can be viewed as being composed of three activities~\cite{Sandholm99}:
\begin{enumerate}
\item \emph{Coalition  structure formation}, corresponding  to the process of  coalitions formation
  such that agents  within each coalition coordinate their activities, but  agents do not coordinate
  between  coalitions. This  means  partitioning the  set  of agents  into  exhaustive and  disjoint
  coalitions. This partition  is called a \emph{coalition structure} (CS).  For instance, given four
  agents  $\{a_1,   a_2,  a_3,   a_4\}$,  a  possible   coalition  structure   is  $\{\{a_1\}\{a_2,
  a_3\},\{a_4\}\}$;
\item \emph{Optimization}:  solving the optimization problem  of each coalition.  This means pooling
  the tasks and resources of the agents in the coalition, and solving this joint problem.
\item \emph{Payoff distribution}: dividing the value of the generated solution among agents.
\end{enumerate}
Even if these activities are independent there  is some interaction. For example, the coalition that
an agent wants to join depends on the portion of the value that the agent would be allocated in each 
potential coalition. This paper focuses on the coalition structure formation in settings where there
are too many coalition structures to enumerate and evaluate due to costly or bounded computation and
limited time.  Instead, agents have  to select a  subset of coalition  structures on which  to focus
their search.  

Specifically, in this paper we propose a stochastic local search procedure named GRASP-CSF to solve the problem of
coalition formation  in CFGs.  The main advantage  of using  a stochastic local  search is  to avoid
exploring an exponential  number of coalition structures providing a  near optimal solution. Indeed,
our  algorithm  does not  provide   guarantees about  finding  the  global  optimal solution.  In
particular the questions we would like to pose are:
\begin{itemize}
\item \textbf{Q1}) can the metaheuristic GRASP be used as a valuable anytime solution for the CSF
problem? In many cases, as in CSF, it is necessary to terminate the algorithm prior to completion due
to time limits and  to reactivity requirements. In this situation, it  is possible to adopt anytime
algorithms, able  to return an  approximation of  the correct answer,  whose quality depends  on the
amount of computation.
\item  \textbf{Q2}) can  the metaheuristic  GRASP be  adopted for  the CSF  problem to  find optimal
  solution faster than the state of the art exact algorithms for CSF problem? In case of optimization
  combinatorial problems, stochastic  local search algorithms have been proved  to be very efficient
  in finding near optimal solution. In many cases, they outperformed the deterministic algorithms in
  computing the optimal solution.
\end{itemize}
The  paper is  organized  as  follows: Section~\ref{sec:definitions}  introduces basic  concepts
regarding  the  CSF problem,  and  Section~\ref{sec:rw} reports  the  related  works about  the
problem. Then, in Section~\ref{sec:grasp} the metaheuristic GRASP applied to the CSF problem will be
presented. Finally,  Section~\ref{sec:exp} will  present the experimental  results of the  proposed algorithm
when compared to some of the state of the art algorithms, and Section~\ref{sec:conc} will conclude the paper.

\section{Definitions}
\label{sec:definitions}

Given a set $A = \{a_1, a_2, \ldots, a_n\}$ of $n$ agents, $|A|=n$, called the \emph{grand coalition}, a
\emph{coalition} $S$ is a non-empty subset of the set $A$, $\emptyset \neq S \subseteq A$.  

A
\emph{coalition structure} $C = \{C_1, C_2, \ldots, C_k \} \subseteq 2^A$ is a partition of the set $A$,
and $k$ is  its size, i.e. $\forall i,j: C_i  \cap C_j = \emptyset$ and $\cup_{i=1}^k  C_i = A$. For
$C = \{C_1, C_2, \ldots, C_k \}$, we define  $\cup C \triangleq \cup_{i=1}^k C_i$. We will denote the set of
all coalition structures of $A$ as $\mathcal M(A)$.  

For instance, given $A =  \{a_1, a_2, a_3\}$,
the possible coalitions are $\{a_1\}$, $\{a_2\}$, $\{a_3\}$, $\{a_1, a_2\}$, $\{a_1, a_3\}$, $\{a_2,
a_3\}$, and $\{a_1, a_2, a_3\}$, while the possible coalition structures are $\{\{a_1\}, \{a_2\},
\{a_3\}\}$, $\{\{a_1\}, \{a_2, a_3\}\}$, $\{\{a_1, a_3\}, \{a_2\}\}$, $\{\{a_1, a_2\}, \{a_3\}\}$, and
$\{\{a_1, a_2, a_3\}\}$.

As  in common  practice, we  consider coalition  formation in  \emph{characteristic  function games}
(CFGs). In CFGs the value of a coalition structure $C$ is given by 
\begin{equation}
  V(C) = \sum_{S \in C} v(S),
\end{equation}
where $v :  2^A \rightarrow \mathbb R$ and  $v(S)$ is the value of the  coalition $S$.  Intuitively,
$v(S)$ represents the maximum payoff the members of $S$ can jointly receive by cooperating.  As 
in~\cite{Sandholm99}, we  assume that  $v(S) \geq  0$. Given  a set of  agents $A$,  the goal  is to
maximize the social welfare of the agents by finding a coalition structure 
\begin{equation}
C^* = \arg max_{C \in \mathcal M(A)} V(C).
\end{equation}

Given $n$ agents, the size of the input to a coalition structure formation algorithm is exponential,
since  it contains the  values $v(\cdot)$  associated to  each of  the $(2^n-1)$  possible coalitions.
Furthermore,  the  number of  coalition  structures  grows as  the  number  of  agents  increases  and
corresponds  to $$\sum_{i=1}^n Z(n,i),$$  where $Z(n,i)$,  also known  as the  Stirling number  of the
second kind, is  the number of coalition structures  with $i$ coalitions, and may  be computed using
the following recurrence: $$Z(n,i) = i Z(n-1,i) + Z(n-1,i-1),$$ where $Z(n,n) = Z(n,1) = 1$. 
As   proved   in~\cite{Sandholm99},  the   number   of   coalition   structures  is   $O(n^n)$   and
$\omega(n^{n/2})$, and hence an exhaustive enumeration becomes prohibitive. 

In this  paper we focus  on games that  are neither \emph{superadditive} nor  \emph{subadditive} for
which  the  problem  of coalition  structure  formation  is  computationally complex.  Indeed,  for
superadditive games  where $v_{S \cup T}  \geq v_S + v_T$  (meaning any two  disjoint coalitions are
better off by merging together), and for subadditive games  where $v_{S \cup T} < v_S + v_T$ for all
disjoint  coalitions  $S,T   \subseteq  A$,  the  problem  of   coalition  structure formation  is
trivial.  In particular,  in  superadditive  games, the  agents  are better  off  forming the  grand
coalition where all agents operate together ($C^*  = \{A\}$), while in subadditive games, the agents
are better off by operating alone ($C^* = \{ \{a_1\}, \{a_2\}, \ldots, \{a_n\}\}$).

\section{Related Work}
\label{sec:rw}
Previous works  on coalition structure formation can  be broadly divided into  two main categories:
exact algorithms that  return an optimal solution,  and approximate algorithms that  find an approximate
solution with limited resources.

A deterministic algorithm must systematically explore the search space of candidate solutions. 
One of the first  algorithm returning an optimal solution is the  dynamic programming algorithm (DP)
proposed in~\cite{Yeh86} for  the set partitioning problem.
This algorithm is  polynomial in the
size of  the input  ($2^n-1$)  and it  runs in $O(3^n)$  time, which is  significantly less  than an
exhaustive enumeration ($O(n^n)$). However, DP is not an anytime algorithm, and has a large memory
requirement.  Indeed, for each coalition $C$ it  computes $t_1(C)$ and $t_2(C)$. It computes all the
possible splits  of the  coalition $C$ and  assign to $t_1(C)$  the best  split and to  $t_2(C)$ its
value. 
In~\cite{Rahwan08} the  authors proposed an  improved version of  the DP algorithm  (IDP) performing
fewer operations and requiring less memory than DP.  IDP, as shown by the authors, is considered one
of the fastest available exact algorithm in the literature computing an optimal solution.

\begin{figure*}[t]
  \centering
  \includegraphics[width = 0.8\textwidth]{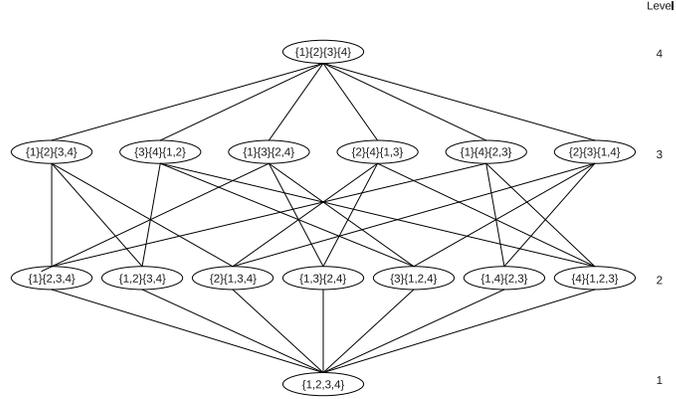}
  \caption{Coalition structure graph for a 4-agent game.}
  \label{fig:coalitiongraph}
\end{figure*}

Both  DP  and IDP  are  not  anytime  algorithms, they  cannot  be  interrupted before  their  normal
termination.   In~\cite{Sandholm99},  Sandholm  et  al.   have   presented  the  first  anytime
algorithm, sketched in Algorithm~\ref{alg:sandholm}, that can  be interrupted to obtain a solution  within a time limit but  not guaranteed to be
optimal.  When not interrupted it returns the optimal solution. 
The coalition structure generation process can be  viewed as a search in a coalition structure graph
as reported in Figure~\ref{fig:coalitiongraph}.
One desideratum is to be able to guarantee that the coalition structure is within a worst case bound 
from optimal, i.e. that searching through a subset $N$ of coalition structures,
\begin{equation}
k = \min \{k' \} \ \mathrm{where} \ k' \geq \frac{V(S^*)}{V(S^*_N)}
\end{equation}
is finite, and as small as possible, where $S^*$ is  the best CS and $S^*_N$ is the best CS that has
been seen in the subset $N$.
In~\cite{Sandholm99} has been proved that:
\begin{enumerate}
\item to bound $k$, it suffices to search the lowest two levels of the coalition structure graph (with
this search, the bound $k=n$, and the number of nodes searched is $2^{n-1}$); 
\item this bound is tight; and, 
\item no other search algorithm can establish any bound $k$ while searching only $2^{n-1}$
nodes or fewer.
\end{enumerate}

\begin{algorithm}
\caption{\textsc{Sandholm et al. algorithm}}
\label{alg:sandholm}
\begin{enumerate}
\item Search the bottom two levels of the coalition structures graph.
\item Continue with a breadth-first search from the top of the graph as long as there is time left,
or until  the entire graph has  been searched (this  occurs when this breadth-first  search completes
level 3 of the graph, i.e. depth n-3).
\item Return the coalition structure that has the highest welfare among those seen so far.
\end{enumerate}
\end{algorithm}

A new anytime algorithm has been proposed in~\cite{Rahwan09}, named IP, whose idea is to partition the space
of the possible solutions into  sub-spaces such that it is possible to compute  upper and lower bounds on
the values of the  best coalition structures they contain. Then, these bounds  are used to prune all
the sub-spaces that cannot contain the  optimal solution. Finally, the algorithm searches through the
remaining sub-spaces  adopting a  branch-and-bound technique avoiding  to examine all  the solutions
within the searched sub-spaces. 

As regards the approximate algorithms, in~\cite{sen00} has been proposed a solution based on a genetic
algorithm, which performs well when there is some regularity in the search space. Indeed, the authors
assume, in  order to  apply their algorithm,  that the value  of a  coalition is dependent  of other
coalitions in the coalition structure, making the algorithm not well suited for the general case.
A  new  solution~\cite{keinanen09} is  based  on  a Simulated  Annealing  algorithm,  a widely  used
stochastic local search method. At each iteration the algorithm  selects a random neighbour solution
$s'$ of a CS $s$. The search proceeds with an adjacent CS $s'$ of the original CS $s$ if $s'$ yields
a better  social welfare than  $s$. Otherwise,  the search is  continued with $s'$  with probability
$e^{(V(s')-V(s))/t}$,  where  $t$  is the  temperature  parameter  that  decreases according  to  the
annealing schedule $t = \alpha t$.

\section{GRASP for CSF}
\label{sec:grasp}
The resource limits posed by MASs, such as the time for finding a solution, require  to have
approximation algorithms  able to quickly  find solutions  that are within  a specific factor  of an
optimal solution.  
In this  section we present  a new anytime  algorithm for CSF based  on a stochastic  local search
procedure, named GRASP-CSF.

A method to find high-quality solutions for a combinatorial problem is a two steps approach consisting of a greedy construction phase followed by a perturbative local search ~\cite{sls04}. Namely, the greedy  construction  method  starts the process from  an  empty candidate  solution and at each  construction step  adds the  best ranked
component  according  to  a   heuristic  selection  function. Successively,  a perturbative  local search algorithm is used  to improve the
candidate  solution  thus  obtained. Advantages  of  this
search  method  are  the  much  better  solution  quality  and  fewer
perturbative  improvement steps  to  reach the  local optimum.  Greedy
Randomized Adaptive  Search Procedures (GRASP)~\cite{feo95}  solve the
problem  of  the  limited  number  of  different  candidate  solutions
generated by  a greedy construction search methods  by randomising the
construction  method.  GRASP  is an  iterative  process,  in  which  each  iteration consists  of  a
construction phase, producing a feasible solution, and a local search 
phase, finding a local optimum  in the neighborhood of the constructed
solution.        The       best       overall        solution       is
returned.

Algorithm~\ref{alg:grasp}  reports   the  outline  for   \textsc{GRASP-CSF}  included  in   the  ELK
system\footnote{ELK is a system  including many algorithms for the CSF problem  whose source code is
  publicly available at \texttt{http://www.di.uniba.it/$\sim$ndm/elk/}.}.  
In  each  iteration,  it
computes a solution $C$ by using a randomised constructive search procedure and then applies a local
search  procedure to  $C$ yielding  an improved  solution. The  main procedure  is made up of two
components: a constructive phase and a local search phase. 

\begin{algorithm}
\caption{\textsc{GRASP-CSF}}
\label{alg:grasp}
\begin{algorithmic}
\REQUIRE{$V$: the characteristic function; $A$: the set of $n$ agents;
\emph{maxiter}: maximum number of iterations}
\ENSURE{solution $\widehat{C} \in \mathcal M(A)$}
\STATE $\widehat{C} = \emptyset$, $V(\widehat{C}) = -\infty$
\STATE iter $= 0$
\WHILE{iter $<$ maxiter}
  \STATE $\alpha = $ rand(0,1)
  \STATE \emph{/* construction */}
  \STATE $C = \emptyset$; $i = 0$
  \WHILE{$i < n$}
    \STATE $\mathcal C = \{ C' | C' = add(C,A)\}$
    \STATE $\overline{s} = \max \{V(T) | T \in \mathcal C\}$
    \STATE $\underline{s} = \min \{V(T) | T \in \mathcal C\}$
    \STATE RCL $= \{C' \in \mathcal C | V(C') \geq \underline{s} + \alpha (\overline{s} - \underline{s})\}$
    \STATE select $T$, at random, from RCL
    \STATE $C$ = $T$
    \STATE $i \leftarrow i + 1$
  \ENDWHILE
  \STATE \emph{/* local search */}
  \STATE $\mathcal N = \{ C' \in neigh(C) | V(C') > V(C)\}$
  \WHILE{$\mathcal N \neq \emptyset$}
    \STATE select $C \in \mathcal N$
    \STATE $\mathcal N \leftarrow \{ C' \in neigh(C) | V(C') > V(C)\}$
  \ENDWHILE
  \IF{$V(C) > V(\widehat{C})$}
  \STATE $\widehat{C} = C$
  \ENDIF
  \STATE iter = iter + 1
\ENDWHILE
\STATE \textbf{return} $\widehat{C}$
\end{algorithmic}
\end{algorithm}

The constructive search algorithm used in GRASP-CSF iteratively adds a solution component by randomly
selecting it,  according to  a uniform  distribution, from a  set, named  \emph{restricted candidate
  list}  (RCL), of  highly ranked  solution components  with respect  to a  greedy function  $g  : C
\rightarrow \mathbb R$.  The probabilistic component  of GRASP-CSF is characterized by randomly choosing
one of  the best candidates  in the  RCL.  In our  case the greedy  function $g$ corresponds  to the
characteristic function $V$ presented in Section~\ref{sec:definitions}.  In particular, given $V$, the heuristic function, and $\mathcal C$, the set of feasible
solution components, $$\underline{s} = \min \{ V(C) | C \in \mathcal C\}$$ and $$\overline{s} = \max \{
V(C)  | C  \in \mathcal  C\}$$ are  computed. Then  the RCL  is defined  by including  in it  all the
components $C$  such that  $$V(C) \geq  \underline{s} + \alpha  (\overline{s} -  \underline{s}).$$ The
parameter  $\alpha$  controls the  amounts  of  greediness and  randomness.  A  value  $\alpha =  1$
corresponds to a greedy construction procedure, while $\alpha = 0$ produces a random construction. 
As  reported  in~\cite{Mockus97},  GRASP  with  a  fixed  nonzero  RCL  parameter  $\alpha$  is  not
asymptotically convergent  to a global optimum.   The solution to make  the algorithm asymptotically
globally convergent, could be to randomly  select the parameter value from the continuous interval
$[0, 1]$ at the beginning of each iteration  and using this value during the entire iteration, as we
implemented in GRASP-CSF.

To improve  the solution generated by  the construction phase, a  local search is used.  It works by
iteratively replacing the current solution with a better solution taken from the neighborhood of the
current solution while there is a better solution in the neighborhood.

In order  to build  the neighborhood of  a coalition  structure $C$, \emph{neigh(C)},  the following
operators, useful to  transform partitions of the grand  coalition, have been used. Given a  CS $C =
\{C_1, C_2, \ldots, C_t\}$: 
\begin{description}
\item[\textbf{split}:] $C \rightarrow C \setminus \{C_i\} \cup \{C_k,C_h\}$\\ where
  $C_k \cup C_h = C_i$, with $C_k, C_h \neq \emptyset$;
\item[\textbf{merge}:]  $C \rightarrow C  \setminus \{C_i,C_j\}_{i \neq
    j} \cup \{C_k\}$\\ where $C_k = C_i \cup C_j$;
\item[\textbf{shift}:] $C \rightarrow C  \setminus \{C_i,C_j\}_{i \neq
    j} \cup \{C_i',  C_j'\}$\\ where $C_i' = C_i  \setminus \{a_i\}$ and $C_j' = C_j  \cup
  \{a_i\}$, with $a_i \in C_i$.
\item[\textbf{exchange}:] $C \rightarrow C  \setminus \{C_i,C_j\}_{i \neq
    j}  \cup \{C_i', C_j'\}$\\  where $C_i'  = C_i  \setminus \{a_i\}  \cup \{a_j\}$  and $C_j'  = C_j
  \setminus \{a_j\} \cup \{a_i\}$, with $a_i \in C_i$ and $a_j \in C_j$;
\item[\textbf{extract}:] $C \rightarrow C \setminus \{C_i\}_{i \neq
    j} \cup \{C_i',  C_j\}$\\ where $C_i' = C_i  \setminus \{a_i\}$ and $C_j =  \{a_i\}$, with $a_i \in C_i$.
\end{description}

In particular, given a set of nonempty subsets of the set of $n$ agents $A$, $C = \{C_1, C_2,
\ldots,  C_t\}$, such  that $C_i  \cap C_j  \neq \emptyset$  and $\cup  C \subset  A$,  the function
$add(C,A)$ used in the construction phase returns a refinement $C'$ obtained from $C$ using
one of the following operators
\begin{enumerate}
\item  $C' \rightarrow C \setminus \{C_i\} \cup
\{C'_i\}$ where $C'_i = C_i \cup \{a_i\}$ and $a_i \not\in \cup C$, or 
\item $C' \rightarrow C \cup
\{C_i\}$ where $C_i = \{a_i\}$ and $a_i \not\in \cup C$.
\end{enumerate}
Starting from the empty  set, in the first iteration all the  coalitions containing exactly one agent
are  considered and  the best  is selected  for further  specialization. At  the iteration  $i$, the
working set of coalition $C$ is refined by trying to add an agent to one of the coalitions in $C$ or
a new coalition containing the new agent is added to $C$.

In the local  search phase, the neighborhood of  a coalition structure $C$ is built  by applying all
the previous operators (split, merge, shift, exchange and extract) to $C$.  As an example, in Table~\ref{tab:tab2} is reported the application of the
operators to the CS $[12][3][4]$. The  problem in using more than the two classical merge
and split operators  corresponds to the fact of  obtaining repetitions of the same  CS. This problem
deserves further attention, each operator must take into account how other operators works. 

\begin{table}
\centering
\begin{tabular}{|c|c|c|c|c|}
\hline
split & merge & shift & exchange & extract\\ 
$[1][2][3][4]$ & $[123][4]$ & $[2][13][4]$ & $[23][1][4]$ & $[1][2][3][4]$\\
& $[124][3]$ & $[2][3][14]$ & $[24][3][1]$ &\\
& $[12][34]$ & $[1][23][4]$ & $[13][2][4]$ &\\
& & $[1][3][24]$ & $[14][3][2]$ &\\
\hline
\end{tabular}
\caption{Operators applied to the CS $[12][3][4]$.}
\label{tab:tab2}
\end{table}


Concerning the representation of the characteristic function and the search space, 
given $n$ agents $A = \{a_1,a_2,\ldots,a_n\}$, we recall that the number of possible coalitions is $2^n-1$.  Hence, the
characteristic function $v : 2^n \rightarrow \mathbb R$ is represented as a vector $CF$ in the following
way. Each subset $S \subseteq A$ (coalition) is described as a binary number $c_{B} = [ b_1, b_2,
\ldots, b_n ]$ where each $b_i = 1$ if $a_i \in S$, $b_i = 0$ otherwise. For instance, given $n=4$,
the coalition $\{a_2,a_3\}$  corresponds to the binary number $0110$. Now,  given the binary representation
of a coalition $S$, its decimal value corresponds to the index in the vector $CF$ where 
its corresponding value $v(S)$ is memorised. This gives us the possibility to have a random access to the values
of  the characteristic  functions in  order to  efficiently  compute the  value $V$  of a  coalition
structure. 

A coalition structure $C = \{C_1, C_2, \ldots, C_k\}$ is represented as an
integer  vector  $cs_D  =  [d_1,  d_2,  \ldots,  d_n]$ where  $\forall  i=1,\ldots,n$:  $d_i  =  j$,
representing that  the agent  $a_i$ belongs  to the  coalition $C_j$, with  $1 \leq  j \leq  k$. For
instance, given $n=4$, the coalition structure  $C = \{C_1, C_2, C_3\} = \{\{1,2\},\{3\},\{4\}\}$ is
represented by the 
vector  $cs_D=[1,1,2,3]$.    Now  in   order  to  compute   $V(C)$,  we   have  to  solve   the  sum
$v(C_1)+v(C_2)+v(C_3)$, where  $C_1$ corresponds to the  binary number $1100$,  $C_2$ corresponds to
the  binary number  $0010$, and  $C_3$ corresponds  to  the binary  number $0001$.   Hence, $V(C)  =
v(C_1)+v(C_2)+v(C_3) =  CF[12]+CF[2]+CF[1]$, where $CF$ is  the vector containing the  values of the
characteristic function. 

\section{Experimental results}
\label{sec:exp}
In order to evaluate our GRASP-CSF, we implemented it in the C language and the corresponding source code has been
included  in   the  ELK  system,   including  some  algorithms   for  CSF,  publicly   available  at
\texttt{http://www.di.uniba.it/$\sim$ndm/elk/}.  
We also implemented the algorithm proposed by Sandholm et al. in \cite{Sandholm99}, DP~\cite{Yeh86}, and IDP~\cite{Rahwan08}, whose source code has been included
in the ELK system. GRASP-CSF has been compared to those algorithms and to the Simulated Annealing
algorithm (SA) proposed in \cite{keinanen09}, kindly provided by the authors. 

In the  following we  present experimental  results on the  behaviour of  these algorithms  for some
benchmarks considering solution qualities and the runtime performances. We firstly compared GRASP-CSF
to DP and IDP. Then we evaluated its ability in generating solutions anytime when compared to the SA
and to the Sandholm et al. algorithms.  

Instances  of the  coalition structure  formation  problem have  been defined  using the  following
distributions for the values of the characteristic function: 
\begin{enumerate}
\item Normal if $v(C) \sim
N(\mu,\sigma^2)$ where $\mu =1$ and $\sigma=0.1$; 
\item  Uniform if $v(C) \sim U(a,b)$ where $a=0$ and
$b=1$.
\end{enumerate}

The algorithms  are executed  on PC with  an Intel  Core2 Duo CPU  T7250 @ 2.00GHz  and 2GB  of RAM,
running Ubuntu kernel 2.6.31.
\subsection{Optimal solution}
\begin{figure}[!h]
  \centering
  \includegraphics[width = 0.5\textwidth]{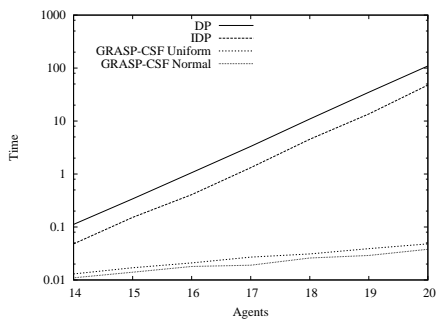}
  \caption{Comparison between DP, IDP and GRASP-CSF.}
  \label{fig:grasp-sa}
\end{figure}

Given different  numbers of  agents, ranging  from 14  to 20, we  compared GRASP-CSF  to DP  and IDP
reporting  the   time  required   to  find   the  optimal  coalition   structure.  As   reported  in
Figure~\ref{fig:grasp-sa},  where  the  time  in  seconds  is plotted  in  a  log  scale,  GRASP-CSF
outperforms both DP and IDP for all  the distributions.  Note that there is one line for DP and IDP
since they do not depend on the input distribution but only on the input dimension. 
Over the  seven problems, the execution  time for DP ranges  from 0.11 seconds (14  agents) to 109.6
seconds (20 agents), for IDP ranges from 0.05 seconds to 47.8 seconds, and for GRASP-CSF ranges from
0.013 seconds  to 0.048 seconds  on average. In  particular, for 20  agents GRASP-CSF is  1000 times
faster  than  IDP  (i.e.  it  takes  0.09\%  of  the  time  taken  by  IDP).  As  we  can  see  from
Figure~\ref{fig:grasp-sa} this  improvement grows  as the  dimension of the  problem grows.  Even if
we cannot make a direct comparison to IP,  as the authors reported in~\cite{Rahwan09}, IP is 570 times
better than IDP in  the case of uniform distribution for 27 agents. Hence,  we can easily argue that
GRASP-CSF is faster than IP.

 As regards
GRASP-CSF we set  the maxiter parameter to $20$  even if in many cases the  optimal coalition structure
has been  found with fewer iterations, see the details in Table~\ref{tab:table1}.  However, this
limit guarantees  to find  always the  optimal coalition. Given  the number  of agents,  10 different
instances of the problem for each distribution have  been generated and time is averaged over the 10
executions. 

Table~\ref{tab:table1} reports an insight of the  GRASP-CSF execution for three problems (18, 19 and
20 agents). The first column reports the  iteration number. For each iteration the table reports the
time and the relative quality of the solution  averaged over 10 instances. The relative quality of a
coalition structure $C$ is obtained as $V(C)/V(C^*)$ where $C^*$ is the optimal coalition structure. As we can see, the quality of the
obtained coalition structures is very high just in the first iterations computed with few milliseconds.

\begin{table}
\centering
\begin{tabular}{|r|cc|cc|cc|}
\hline
& \multicolumn{2}{c|}{18 agents} & \multicolumn{2}{c|}{19 agents} & \multicolumn{2}{c|}{20 agents}\\
I & Time & Quality  & Time & Quality  & Time & Quality \\
\hline \hline
1&0.0015&	0.9767&	0.0019&	0.9866&	0.0031&	0.9703\\
2&0.0036&	0.9941&	0.0032&	0.9919&	0.0055&	0.9928\\
3&0.0052&	0.9910&	0.0062&	0.9980&	0.0070&	0.9973\\
4&0.0059&	0.9930&	0.0085&	0.9993&	0.0104&	0.9973\\
5&0.0078&	0.9976&	0.0085&	0.9976&	0.0132&	0.9957\\
6&0.0094&	0.9942&	0.0121&	0.9973&	0.0137&	0.9973\\
7&0.0106&	1.0000&	0.0137&	0.9992&	0.0171&	0.9937\\
8&0.0127&	1.0000&	0.0157&	0.9935&	0.0207&	0.9944\\
9&0.0153&	0.9962&	0.0183&	0.9992&	0.0227&	0.9973\\
10&0.0142&	0.9987&	0.0190&	1.0000&	0.0244&	0.9973\\
11&0.0175&	1.0000&	0.0211&	0.9996&	0.0280&	0.9977\\
12&0.0193&	0.9976&	0.0236&	0.9996&	0.0303&	1.0000\\
13&0.0200&	1.0000&	0.0249&	1.0000&	0.0319&	1.0000\\
14&0.0215&	0.9962&	0.0261&	0.9996&	0.0337&	0.9981\\
15&0.0240&	1.0000&	0.0285&	1.0000&	0.0384&	0.9973\\
16&0.0259&	1.0000&	0.0325&	1.0000&	0.0380&	1.0000\\
17&0.0281&	1.0000&	0.0347&	1.0000&	0.0430&	1.0000\\
18&0.0309&	1.0000&	0.0370&	1.0000&	0.0448&	0.9980\\
19&0.0299&	1.0000&	0.0370&	1.0000&	0.0451&	0.9981\\
20&0.0297& 1.0000& 0.0399& 1.0000& 0.0499& 1.0000\\
\hline
\end{tabular}
\caption{Time (in seconds) and relative quality of GRASP-CSF obtained on the first 20 iterations.}
\label{tab:table1}
\end{table}



\subsection{Approximate solution}
\begin{figure}[!h]
  \centering
  \includegraphics[width = 0.5\textwidth]{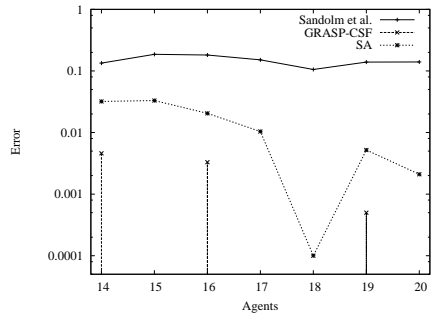}
  \caption{Comparison of Sandholm et al., Simulated Annealing (SA) and GRASP-CSF.}
  \label{fig:anytime}
\end{figure}

In  this second  experiment  we  compared the  anytime  characteristic of  GRASP-CSF  to  the Sandholm  et
al.  algorithm~\cite{Sandholm99}   and  the  Simulated   Annealing  algorithm~\cite{keinanen09}.  We
generated 10 instances for  each problem with agents ranging from 14  to 20 and uniform distribution
$U(0,1)$.  For each problem  we set  a time  limit to  return a  good solution  and we  recorded the
relative error of the obtained solution $S$ by each of the three algorithms computed as $e =
1-V(S)/V(S^*)$, where $S^*$ is the best coalition structure. Figure~\ref{fig:anytime} plots the error in log scale
averaged over the  10 instances for each problem. As  we can see GRASP-CSF is always  able to find a
better coalition structure than those obtained by Sandholm et al. and SA. With this second experiment we can conclude
that GRASP-CSF quickly finds very good solutions.




\section{Conclusions}
\label{sec:conc}
The paper presented the application of the stochastic local search GRASP to the problem of coalition
structure formation.   As reported  in the experimental  section the proposed  algorithm outperforms
some of the state of the art algorithms in computing optimal coalition structures. 

As a  future work it  should be interesting  to investigate the behaviour  of the operators  used to
create the neighborhood of  a coalition structure. In particular, an in  deep study may be conducted
in  learning  to  choose  the  correct  operators  respect to  the  distribution  of  the  coalition
values. Furthermore, as already previously discussed, the application of shift, exchange and extract
operators should  generate repetitions of the same  coalition structure obtained with  the split and
merge operators.  Hence, an analysis  on how to  overcome this problem,  avoiding to spend  time and
space resources, deserves more attention.

\bibliographystyle{splncs}
\bibliography{coalition}

\end{document}